\documentclass[letterpaper]{article}
\usepackage{aaai}
\usepackage{times}
\usepackage{helvet}
\usepackage{amsmath}
\usepackage{graphicx}
\usepackage{subfig}
\usepackage[noend] {algorithm2e, algpseudocode}
\usepackage{courier}
\DeclareMathOperator*{\argmax}{arg max}
\begin{document}
%

\RestyleAlgo{boxruled}
\LinesNumbered

\title{Principal Component Properties of Adversarial Samples}

\author{Malhar Jere\textsuperscript{1},
Sandro Herbig \textsuperscript{2},
Christine Lind\textsuperscript{1}, 
Farinaz Koushanfar \textsuperscript{1}\\
\textsuperscript{1}{University of California San Diego}\\
\textsuperscript{2}{University of Erlangen-Nuremberg, Erlangen, Germany}
}
\maketitle
\begin{abstract}
\begin{quote}
Deep Neural Networks for image classification have been found to be vulnerable to adversarial samples, which consist of sub-perceptual noise added to a benign image that can easily fool trained neural networks, posing a significant risk to their commercial deployment. In this work, we analyze adversarial samples through the lens of their contributions to the principal components of \textit{each} image, which is different than prior works in which authors performed PCA on the entire dataset. We investigate a number of state-of-the-art deep neural networks trained on ImageNet 
as well as several attacks for each of the networks. Our results demonstrate empirically that adversarial samples across several attacks have similar properties in their contributions to the principal components of neural network inputs. We propose a new metric for neural networks to measure their robustness to adversarial samples, termed the $(k,p)$ point. We utilize this metric to achieve 93.36\% accuracy in detecting adversarial samples independent of architecture and attack type for models trained on ImageNet.

\end{quote}
\end{abstract}

\section{Introduction}
\noindent Artificial Neural Networks have made a resurgence in recent times and have achieved state of the art results on numerous tasks such as image classification \cite{russakovsky2015imagenet}. As their popularity rises the investigation of their security will become ever more relevant. Adversarial examples in particular - which involve small, tailored changes to the input to make the neural network misclassify it - pose a serious threat to the safe utilization of neural networks. Recent works have shown that adversarial samples comprise of \textit{non-robust} features of datasets, and that neural networks trained on adversarial samples can generalize to the test set \cite{ilyas2019adversarial}. Because these non-robust features are invisible for humans, performing inference on lossy reconstructions of the adversarial input has the potential to shed light on the dependence between the adversarial noise and the robust features of the image.


In this work, we seek to analyze adversarial samples in terms of their contribution to the principal components of an image and characterize the vulnerability of these models. We test our method for a number of different Deep Neural Network architectures, datasets and attack types, and identify a general trend about adversarial samples.

\section{Background and Prior Work}
\subsection{Adversarial Samples}
\noindent We consider a neural network $f( \cdot)$ used for classification where $f(x)_{i}$ represents the probability that image $x$ corresponds to class $i$. Images are represented as $x \in [0,1]^{w.h.c}$, where $w, h, c$ are the width, height and number of channels of the image. We denote the classification of the network as $c(x) = \argmax_{i} f(x)_{i}$, with $c^{\ast}(x)$ representing the true class, or the ground truth of the image.
Given an image $x$ and an image classifier $f(\cdot)$, an adversarial sample $x'$ follows two properties:
\begin{itemize}
    \item $D(x,x')$ is small for some distance metric $D$, implying that the images $x$ and $x'$ appear visually similar to humans.
    \item $c(x') \neq c^{\ast}(x) = c(x)$. This means that the prediction on the adversarial sample is incorrect whereas the original prediction is correct.
\end{itemize}

\noindent In this work, we focus on 3 methods to generate adversarial samples.

\subsubsection{DeepFool.} Deepfool \cite{moosavi2016deepfool} is an iterative untargeted attack technique to manipulate the decision boundaries of neural networks while minimizing the $L_2$ distance metric between the altered (adversarial) example and the original image. 

\subsubsection{Jacobian Saliency Map Attack:}
Papernot et al. introduced the Jacobian-based Saliency Map Attack \cite{papernot2016limitations}, a targeted attack optimized under the $L_{0}$ distance. The attack is a greedy algorithm that utilizes the \textit{saliency map} of neural networks to pick pixels to modify one at a time, increasing the target classification on each iteration. 

\subsubsection{Carlini Wagner Attack.}
For a given image, the goal of the Carlini Wagner attack \cite{carlini2017adversarial} is to find a small perturbation such that the model misclassifies the input as a chosen adversarial class. 
The attack can be formulated as the following optimization problem: 
\ $min {|| \delta ||}_p + c \cdot f(x + \delta) $ \ such that \ $ x + \delta \in [0,1]^{n}$\, where ${|| \delta ||}_p$ is the p-norm. In this paper we use the $L_{2}$ norm i.e. $p=2$. 


\subsection{Prior Work}
There have been several prior works in detecting adversarial samples. DeepFense\cite{rouhani2018deepfense} formalizes the goal of thwarting adversarial attacks as an optimization problem that minimizes the rarely observed regions in the latent feature space spanned by a neural network.  \cite{pang2018towards} seek to minimize the reverse cross-entropy which encourage deep networks to learn latent representations that better distinguish adversarial examples from normal ones. \cite{ma2019nic} identify exploitation channels and utilize them for adversarial sample detection. 

Our work is most similar to \cite{ma2018characterizing}, which characterizes adversarial samples in terms of the Local Intrinsic Dimensionality, and to \cite{carlini2017adversarial} and \cite{bhagoji2017dimensionality}, which show PCA to be an effective defense against certain adversarial attacks on smaller datasets such as MNIST. Our method, however, is different in that we seek to understand adversarial samples based on their contributions to the principal components of a \textit{single} image, and that we use the rows as principal components, thereby allowing us to scale our technique to much larger datasets such as ImageNet. 

\section{Methodology}
\subsection{Threat Model} There are two different settings for adversarial attacks. The most general setting is the black box threat model where adversaries do not have access to any information about the neural network (e.g. gradient) except for the predictions of the network. In the white box threat model all information about the neural network is accessible, including its weights, architecture, gradients and training method. In this work we consider situations where adversaries have white-box access to the neural network.

\subsection{Defensive PCA} 
\cite{carlini2017adversarial} and \cite{bhagoji2017dimensionality} have shown PCA to be an effective defense against certain adversarial attacks on smaller datasets, where $n$ is the number of samples in the dataset and $d$ the number of features $(rows \times columns)$ of each sample. This works well when the dataset and number of features are small, however, for larger datasets with larger inputs this method becomes computationally inefficient as the size of the data matrix scales quadratically with the size of dataset.

To tackle this emerging problem we suggest an alternative way to perform PCA, where $n=w$ is the number of rows and $d = h\times c$ is the product of the number of columns and the channels of an image $x \in [0,1]^{w.h.c}$. In doing so we can capture the correlations between pixels of an image and vastly reduce the number of dimensions required for PCA. Additionally, this method is independent of the dataset size.  Furthermore, our method also has the added advantage that it requires no knowledge of the dataset which makes it more versatile.

\begin{figure}[ht!]
    \centering
    \includegraphics[scale=0.48]{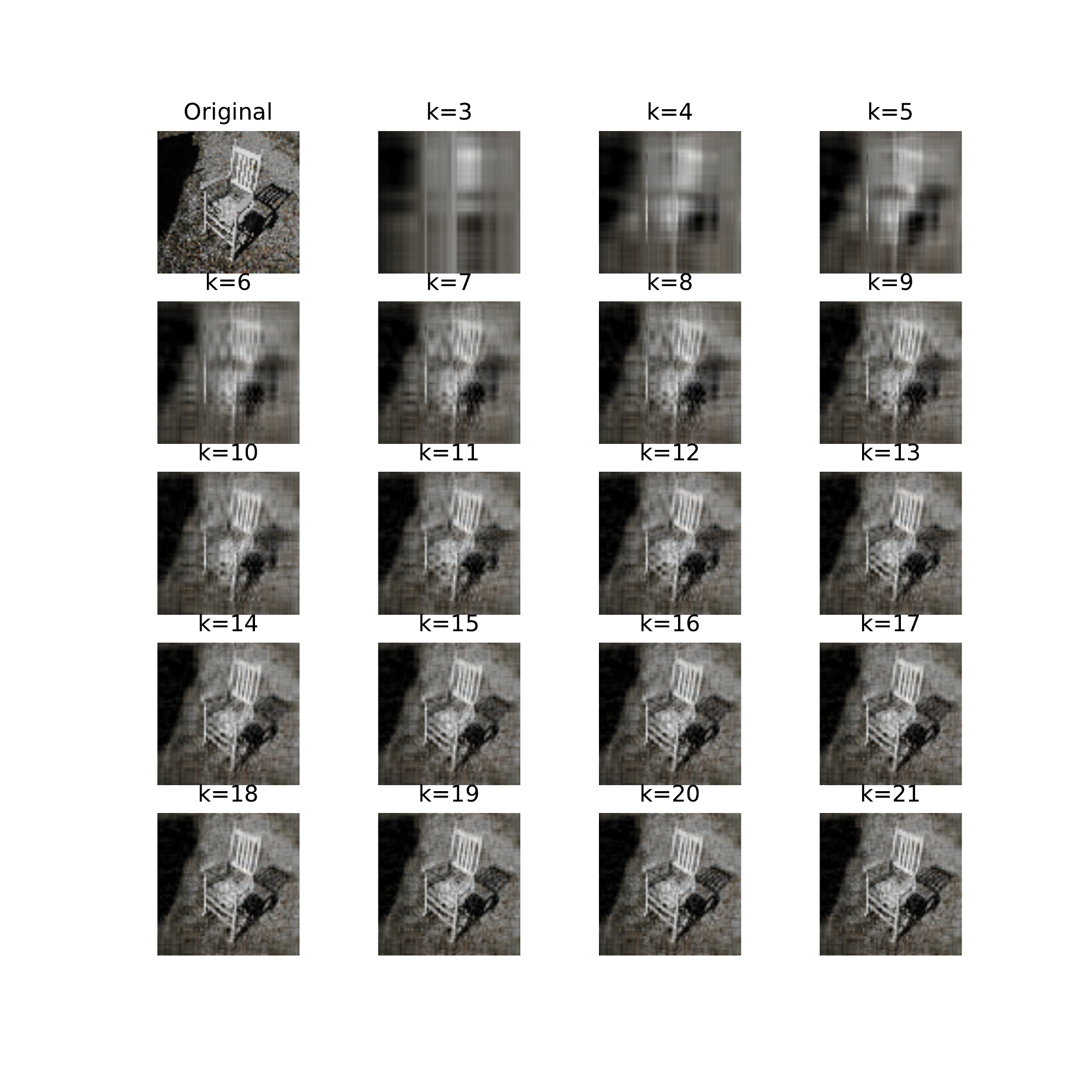}
    \caption{Examples of PCA reconstructed images for a randomly chosen image from the ImageNet validation dataset.}
    \label{pca_examples}
\end{figure}

We term this new method of performing PCA as \textit{rowPCA} denoted as $C = P_{row}(x)$ for an input image $x \in [0,1]^{w.h.c}$, which treats each \textit{$h \times c$} row of $x$ as a principal axis. As an example, an ImageNet input image $x$ with dimensionality $(224\times224\times3)$ will generate $224$ principal components. We can then reconstruct our image $x$ from the principal components with smaller components contributing smaller variance to the image. We denote the first $i$ principal components $[c_1, c_2, ... c_i]$ as $C_{1:i}$, and the image reconstruction operation as $P_{inv, row}(\cdot)$. The reconstructed image $x^{\ast}$ generated from the first $i$ row principal components is thus $x^{\ast} = P_{inv, row}(C_{1:i})$. Figure \ref{pca_examples} shows several reconstructed inputs for a benign sample.

\subsection{Detecting Dominant Classes}
\begin{figure}[ht!]
    \centering
    \includegraphics[scale=0.32]{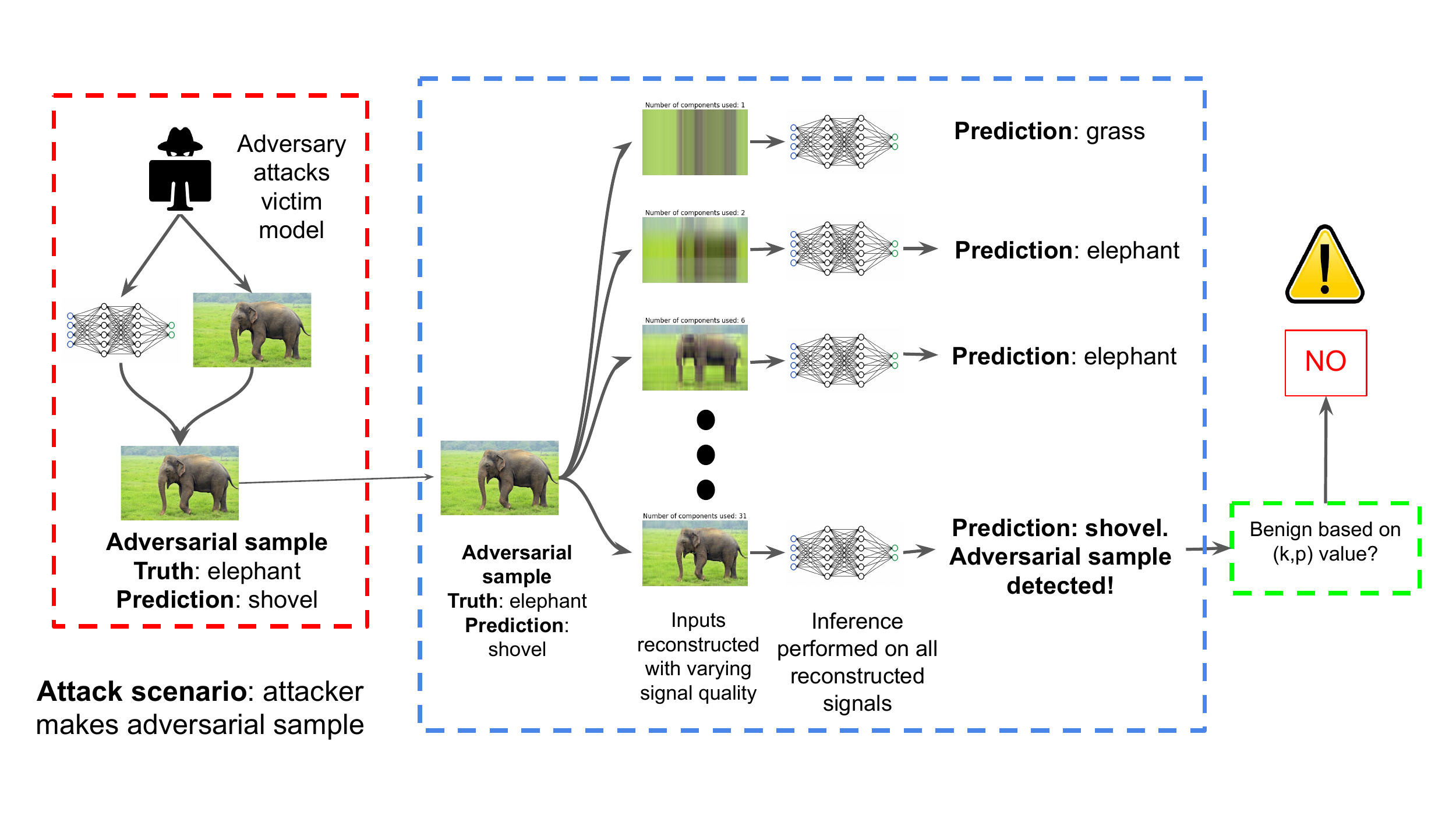}
    \caption{Visualization of defensive PCA applied to an adversarial input. For an input image, we reconstruct the image from the principal components and perform inference on each to determine the component when the dominant class starts becoming the top prediction. The dominant class could be the adversarial class for adversarial inputs, or the ground truth or misclassified class for benign inputs.}
    \label{all_samples_all_architectures}
\end{figure}

We define the dominant class as the predicted class on the full image $x$. The $(k,p)$ point is defined as a tuple consisting of the component when the dominant class starts becoming the top prediction and the softmax probability $p$ of the dominant class at that particular component number. Algorithm \ref{kppoint} outlines the procedure to obtain the $(k,p)$ point for a particular input, and Figure \ref{all_samples_all_architectures} demonstrates the functionality of our detection method on an adversarial sample. The steps that occur are:
\begin{itemize}
    \item The input image is decomposed into its principal components by the rows.
    \item Each of the sets $C_{1:k}$ of descending principal components (sorted by by eigenvalue) is used to reconstruct the image.
    \item Each reconstructed image is fed through the neural network and the predictions are observed.
    \item The $(k,p)$ point is found for the particular set of predictions for each image and is subsequently used to determine whether that particular sample is adversarial or benign.
\end{itemize}

\begin{algorithm}[h!]
\KwResult{ $(k,p)$ point}
\textbf{begin:} \\
 $k = n$; \\
 $topper = argmax f(x)$ \\
 \While{$topper ==  c(x)$}{
  $k = k - 1$\;
  $x^{\ast} = P_{inv, row}(C_{1:k})$ \\
  $topper = argmax f(x^{\ast})$
 }
 $p = p(x^{\ast})[topper]$ \\
 \algorithmicreturn{$(k,p)$}
 \caption{Finding the $(k,p)$ point for a given neural network $f(\cdot)$, input image $x$, top scoring class on input image $c(x)$, and maximum number of principal components $n$. We reconstruct the image from components $1$ through $i$ and find the point at which the dominant class is no longer dominant.}
\label{kppoint}
\end{algorithm}

\section{Experiments}

\subsection{Experimental Setup} 
\noindent \textbf{Datasets and Models:} We evaluated our method on neural networks pre-trained on the ImageNet dataset \cite{deng2009imagenet} in PyTorch, namely Inception-v3, \cite{szegedy2016rethinking}, Resnet-50 \cite{he2016deep},and VGG19 \cite{simonyan2014very}.


\noindent \textbf{Attack methods:} 
For each of the models we evaluated our method on the DeepFool \cite{moosavi2016deepfool}, Jacobian Saliency Map Attack (JSMA) \cite{papernot2016limitations} and Carlini-Wagner L2 attack~\cite{carlini2017adversarial} using the Foolbox library \cite{rauber2017foolbox}. For each of the 9 \textit{(attack, model)} pairs, we generated 100 adversarial images.



\subsection{Results}

\subsubsection{Behavior of adversarial samples.} 
\begin{figure}[ht!]
\centering
\subfloat[$(k,p)$ points for Carlini-Wagner adversarial samples]{
  \includegraphics[scale=0.25]
  {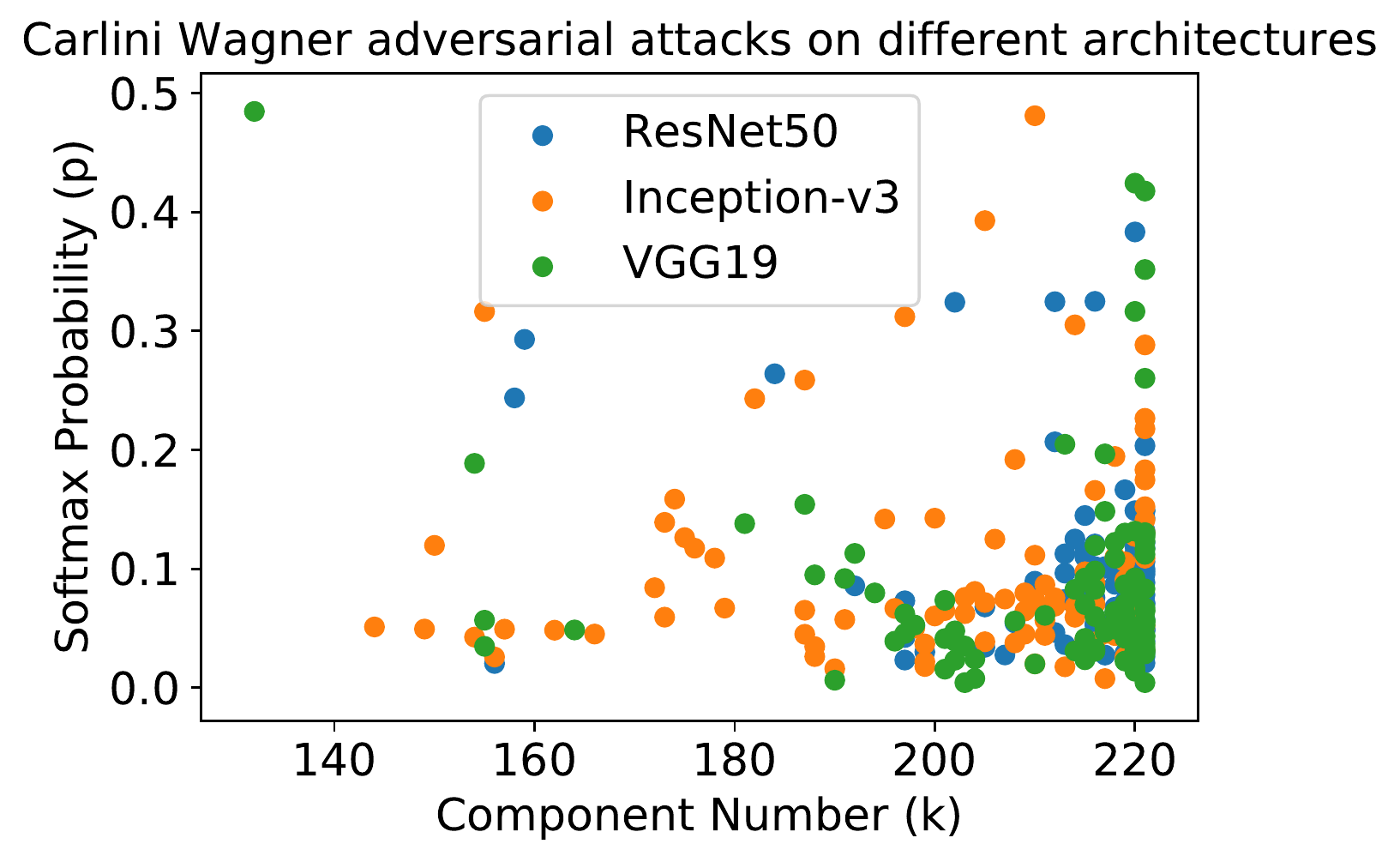}
  \label{CW_imagenet}
}
\subfloat[$(k,p)$ points for DeepFool adversarial samples]{
  \includegraphics[scale=0.25]
  {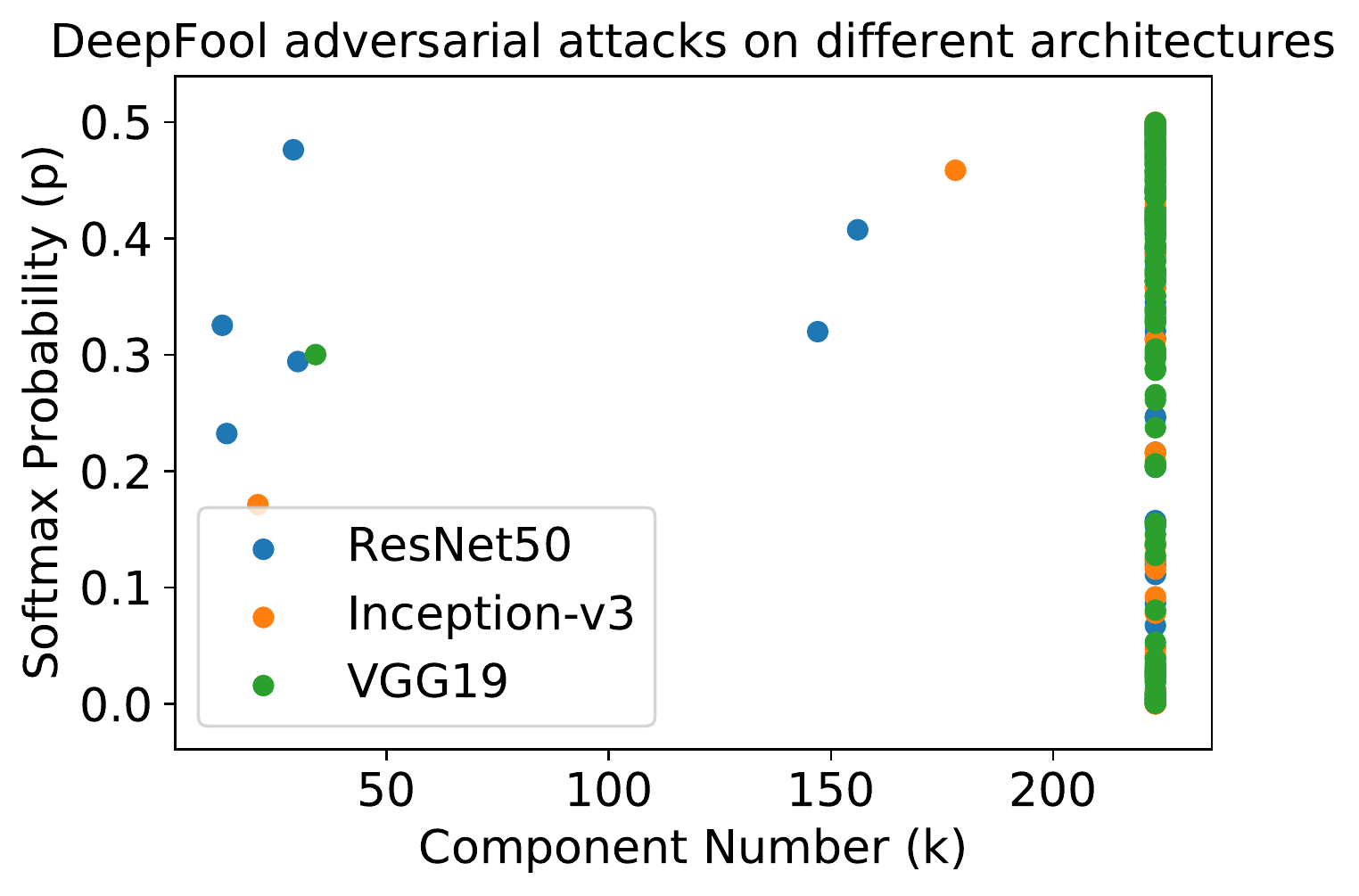}
  \label{df_imagenet}
}
\hspace{0mm}
\subfloat[$(k,p)$ points for JSMA adversarial samples]{
  \includegraphics[scale=0.25]
  {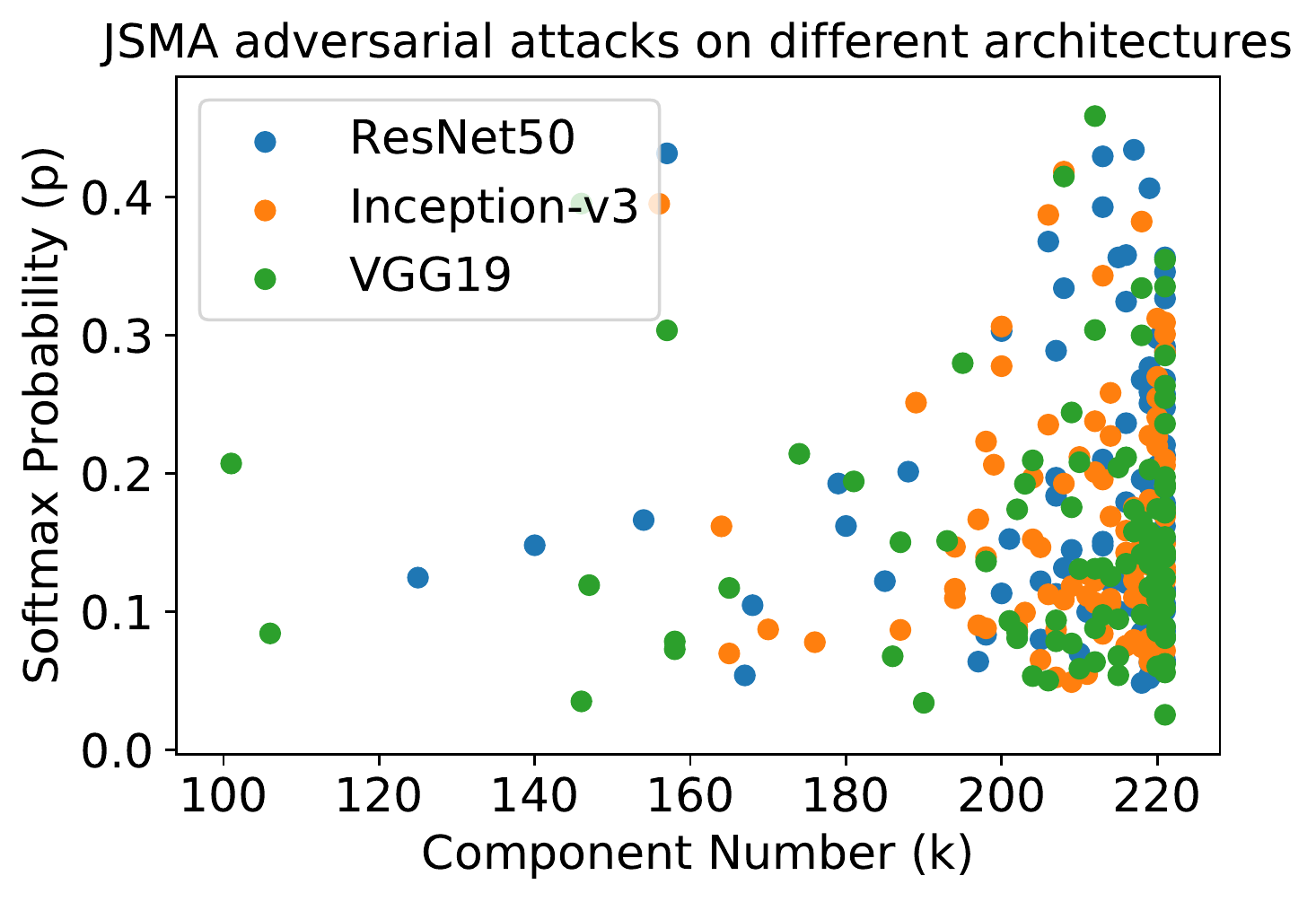}
  \label{jsma_imagenet}
  
}
\subfloat[$(k,p)$ points for benign samples]{
  \includegraphics[scale=0.25]
  {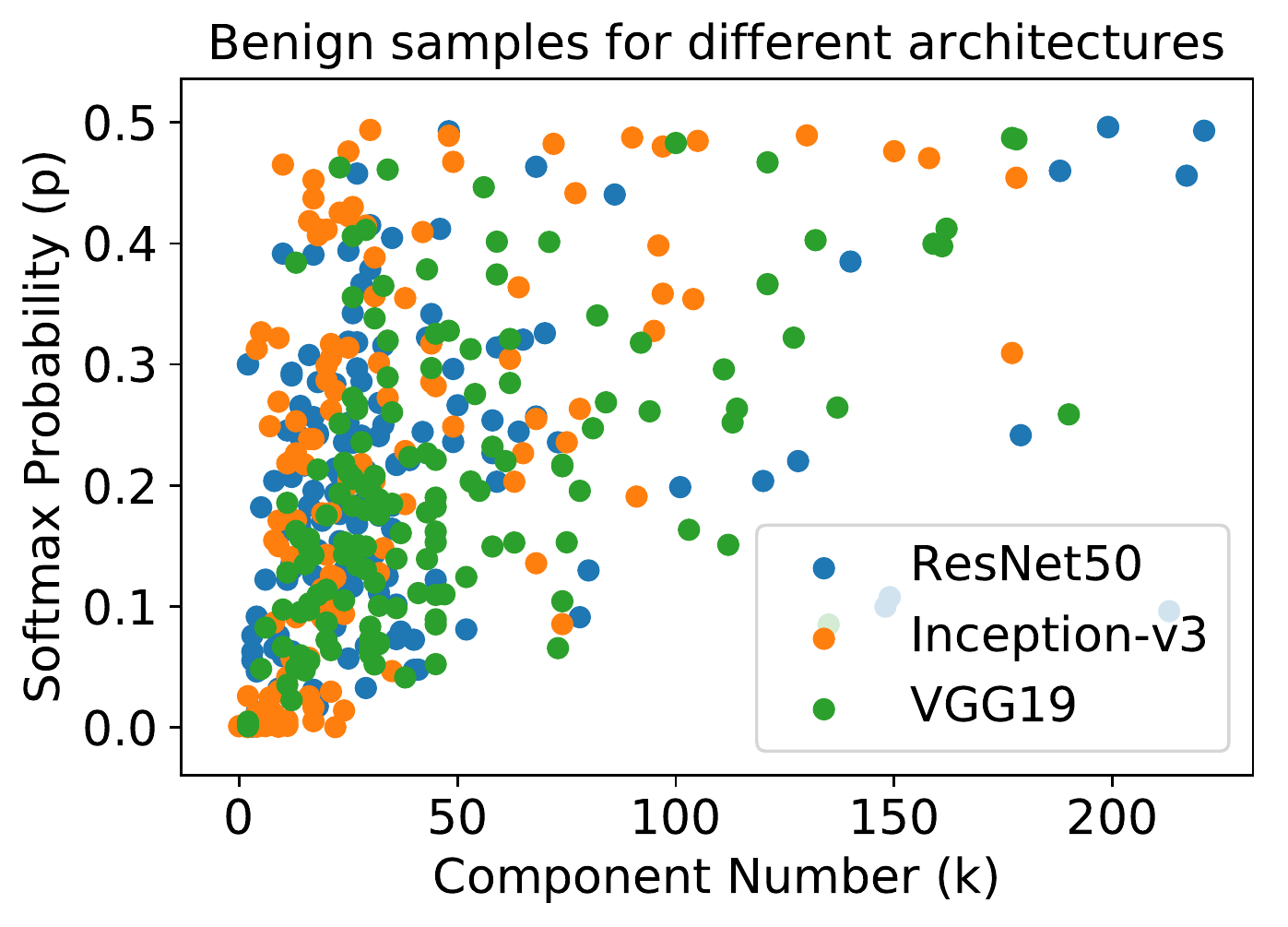}
  \label{benign_imagenet}
}
\caption{$(k,p)$ points for adversarial and benign samples for ImageNet trained models.}
\end{figure}

\begin{figure}[ht!]
    \centering
    \includegraphics[scale=0.35]
    {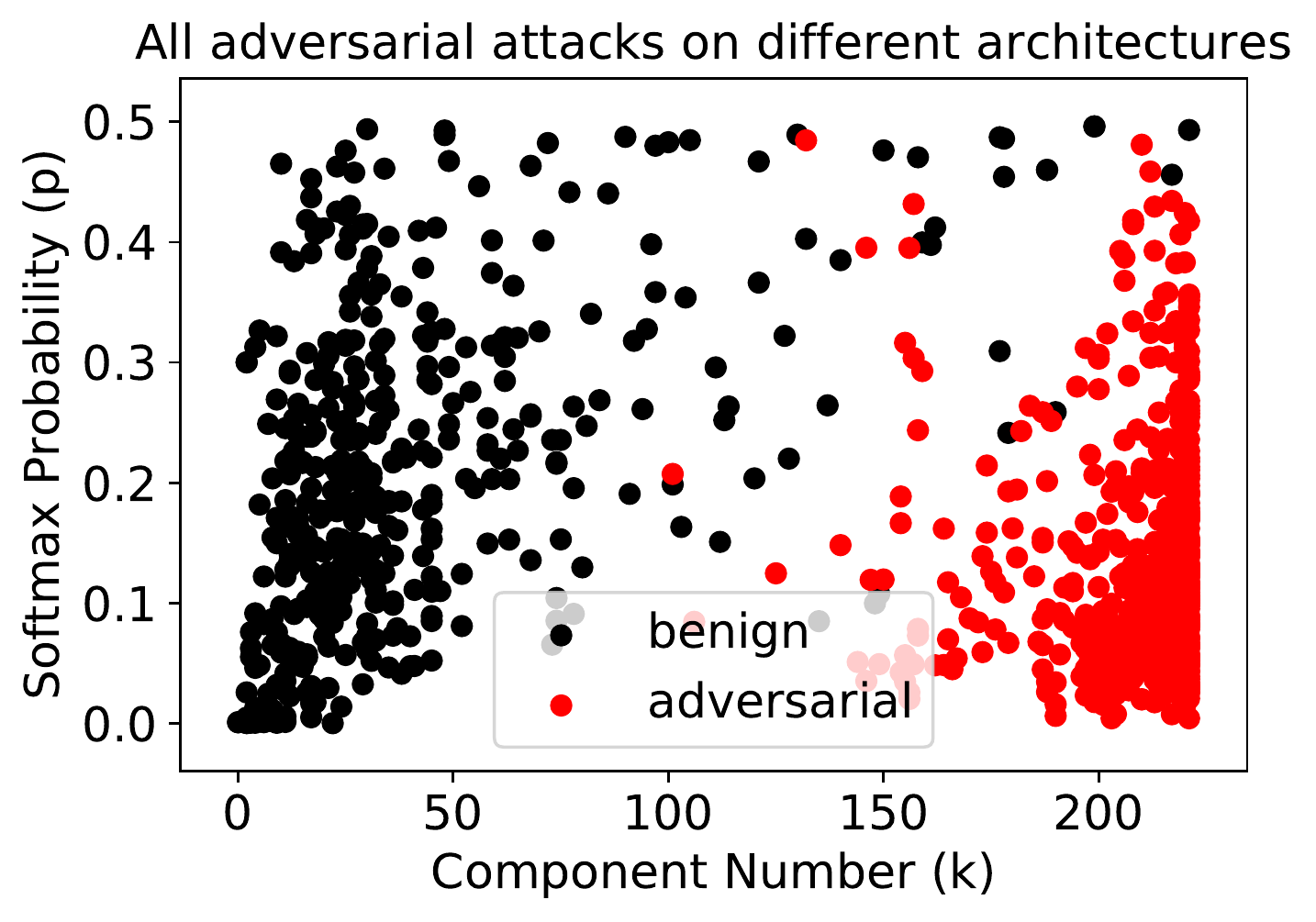}
    \caption{$(k,p)$ points for all benign and adversarial ImageNet samples for all models and adversarial attacks.}
    \label{all_attacks_all_benigns}
\end{figure}

Figures \ref{CW_imagenet}, \ref{df_imagenet} and \ref{jsma_imagenet} shows the clustering of adversarial samples in similar regions of the $(k,p)$ space, while figure \ref{benign_imagenet} shows the clustering of benign samples in similar regions of the $(k,p)$ space. Figure \ref{all_attacks_all_benigns} shows the $(k,p)$ points of all the adversarial and benign points, demonstrating their separability.

\subsubsection{Detection of adversarial samples}  We train binary classifiers on the $(k,p)$ points for a fixed $(attack, model)$ pair and evaluate them against points from the same pair as well as adversarial samples derived using other attacks targeted towards different architectures.

\begin{figure}[ht!]
    \centering
    \includegraphics[scale=0.19]{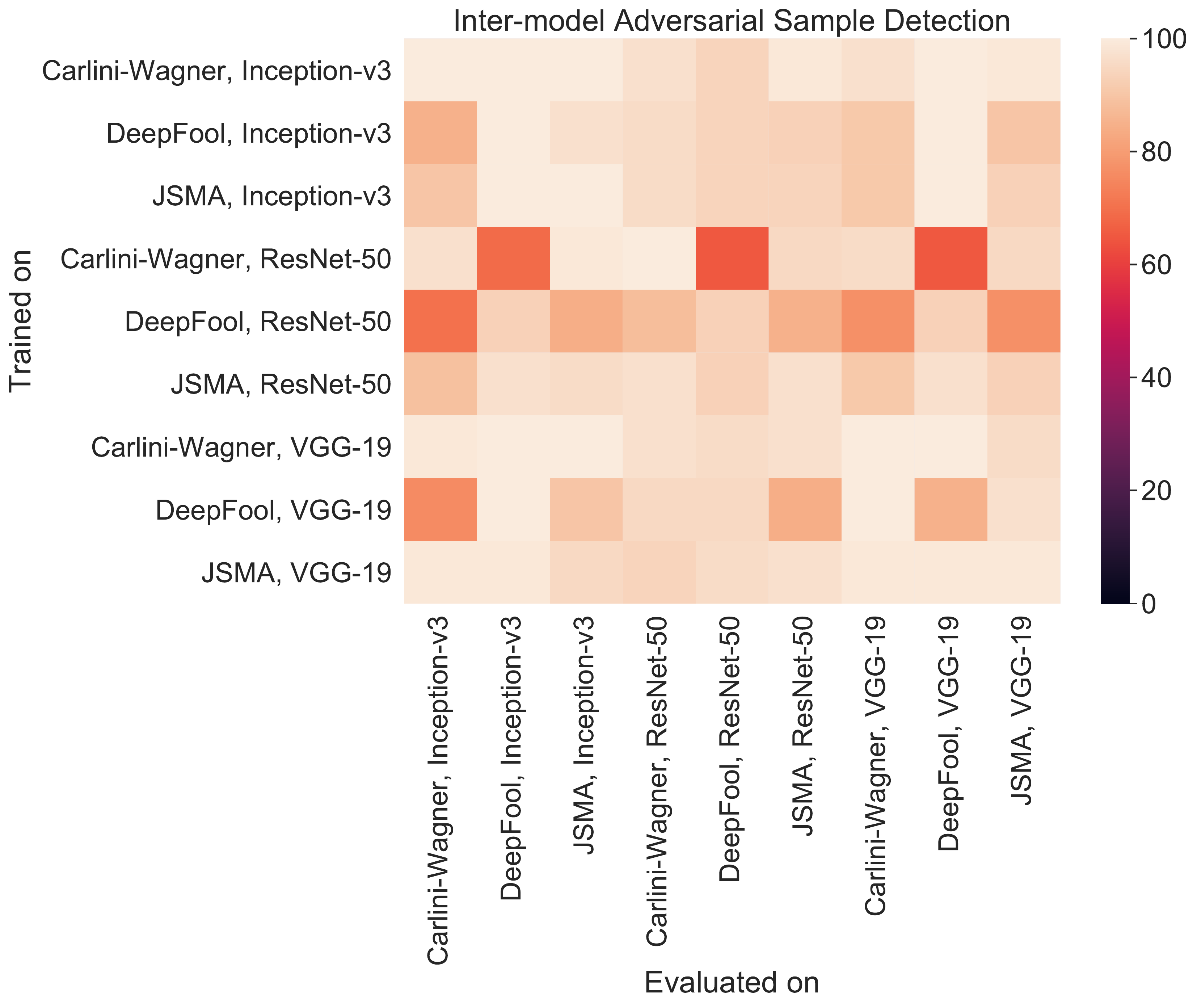}
    \caption{Inter-model and Intra-model adversarial sample detection. We achieve near perfect prediction rates for simple discriminative models trained to identify adversarial samples from one \textit{(attack, model)} pair and evaluated on a different one. The $y$ axis represents the \textit{(attack, model)} we trained our classifier to identify, and the $x$ axis represents the \textit{(attack, model)} we evaluated our classifier on.}
    \label{Inter_model_adversarial_sample_detection}
\end{figure}

\begin{itemize}
    \item \textbf{Intra-model detection rate: } Given a \textit{(attack, model)} pair, this metric measures the probability of predicting whether a given $(k,p)$ point is  either a benign or adversarial sample. We gathered 128 correctly-predicted benign samples and 100 adversarial points for each \textit{(attack, model)} pair for the ImageNet dataset and used an AdaBoost \cite{freund1999short} classifier with 200 weak Decision Tree estimators to distinguish between the two. We achieve an average prediction rate of 94.81\%, namely that we can correctly predict whether a sample for a given neural network will be adversarial or benign 94.81\% of the time.
    
    \item \textbf{Inter-model detection rate: } We observe the $(k,p)$ distributions of adversarial samples across all attack types and models in order to determine their similarity. To measure this, we train classifiers trained on one \textit{(attack, model)} pair and evaluate them on benign and adversarial samples for every other \textit{(attack, model)} pair, as demonstrated in Figure \ref{Inter_model_adversarial_sample_detection}. We achieve an average adversarial detection rate of 93.36\% across all architectures and adversarial methods.
    
\end{itemize}

\section{Discussion}
PCA is one of numerous linear methods for dimensionality reduction of neural network inputs. Other techniques such as Sparse Dictionary Learning and Local Linear Embedding are potential alternatives to PCA, which we intend on exploring in future work. One particular limitation of our method is the need for many rows in the input, which would make our defense inapplicable to inputs for smaller neural networks.

\section{Conclusion}
We identify a new metric, the $(k,p)$ point, to analyze adversarial samples in terms of their contributions to the principal components of an image. We demonstrate empirically that the $(k,p)$ points of benign and adversarial samples are distinguishable across adversarial attacks and neural network architectures and are an underlying property of the dataset itself. We train a binary classifier to detect adversarial samples and achieve a 93.36\% detection success rate.



\bibliography{bibliography}

\end{document}